\documentclass{article}

\usepackage[nonatbib,preprint]{neurips_2020}

\usepackage[utf8]{inputenc} 
\usepackage[T1]{fontenc}    
\usepackage{hyperref}       
\usepackage{url}            
\usepackage{booktabs}       
\usepackage{amsfonts}       
\usepackage{nicefrac}       
\usepackage{microtype}      

\usepackage{enumitem}
\usepackage{multirow}
\usepackage{arydshln}
\usepackage{amsmath}
\usepackage{makecell}
\usepackage{diagbox}
\usepackage{algorithm,algpseudocode}
\algnewcommand\Input{\item[\textbf{Input:}]}%
\algnewcommand\Output{\item[\textbf{Output:}]}%
\algnewcommand{\To}{\textbf{To }}

\usepackage{xcolor}

\newcommand{\norm}[1]{\left\lVert#1\right\rVert}

\usepackage{epsfig}
\usepackage{graphicx}

\usepackage{adjustbox}
\usepackage{array}
\usepackage{booktabs}
\usepackage{colortbl}
\usepackage{float,wrapfig}
\usepackage{hhline}
\usepackage{multirow}

\usepackage{amsmath,amsfonts,amsthm,amssymb}
\usepackage{bm}
\usepackage{nicefrac}
\usepackage{microtype}

\usepackage{changepage}
\usepackage{extramarks}
\usepackage{fancyhdr}
\usepackage{lastpage}
\usepackage{setspace}
\usepackage{soul}
\usepackage{xspace}

\newcommand*{\mathcolor}{}
\def\mathcolor#1#{\mathcoloraux{#1}}
\newcommand*{\mathcoloraux}[3]{%
  \protect\leavevmode
  \begingroup
    \color#1{#2}#3%
  \endgroup
}

\newcolumntype{L}[1]{>{\raggedright\let\newline\\\arraybackslash\hspace{0pt}}m{#1}}
\newcolumntype{C}[1]{>{\centering\let\newline\\\arraybackslash\hspace{0pt}}m{#1}}
\newcolumntype{R}[1]{>{\raggedleft\let\newline\\\arraybackslash\hspace{0pt}}m{#1}}

\newcommand{\ignore}[1]{}

\makeatletter
\DeclareRobustCommand\onedot{\futurelet\@let@token\@onedot}
\def\@onedot{\ifx\@let@token.\else.\null\fi\xspace}

\makeatother

\definecolor{MyDarkBlue}{rgb}{0,0.08,1}
\definecolor{MyDarkGreen}{rgb}{0.02,0.6,0.02}
\definecolor{MyDarkRed}{rgb}{0.8,0.02,0.02}
\definecolor{MyDarkOrange}{rgb}{0.40,0.2,0.02}
\definecolor{MyPurple}{RGB}{111,0,255}
\definecolor{MyRed}{rgb}{1.0,0.0,0.0}
\definecolor{MyGold}{rgb}{0.75,0.6,0.12}
\definecolor{MyDarkgray}{rgb}{0.66, 0.66, 0.66}

\title{Model Compression Using Optimal Transport}

\author{%
 Suhas Lohit \quad \quad Michael Jones \\
  Mitsubishi Electric Research Laboratories\\
  Cambridge, MA \\
  \texttt{slohit@merl.com, mjones@merl.com} \\
}

\begin{document}

\maketitle

\begin{abstract}
Model compression methods are important to allow for easier deployment of deep learning models in compute, memory and energy-constrained environments such as mobile phones. Knowledge distillation is a class of model compression algorithm where knowledge from a large teacher network is transferred to a smaller student network thereby improving the student's performance. In this paper, we show how optimal transport-based loss functions can be used for training a student network which encourages learning student network parameters that help bring the distribution of student features closer to that of the teacher features. We present image classification results on CIFAR-100, SVHN and ImageNet and show that the proposed optimal transport loss functions perform comparably to or better than other loss functions. 
\end{abstract}

\section{Introduction}
Deep convolutional neural networks are currently the most effective methods for many applications in computer vision including image classification and object detection \cite{khan2019survey}. However, network architectures yielding state-of-the-art network accuracies are memory and compute heavy \cite{canziani2016analysis}. Furthermore, recent evidence shows that it is necessary for neural networks to be overparameterized for gradient-based training procedures to be maximally effective \cite{PoggioEtAl2018,Belkin15849,ZhangEtAl2017}. This makes it difficult to employ them in resource-constrained environments such as mobile phones, drones etc. In order to overcome this disadvantage and reduce computational and memory requirements for inference, several model compression techniques have been shown to be of value. In particular, several knowledge distillation methods have been devised in order to improve the accuracy of smaller networks -- student networks -- by transferring the knowledge from a more accurate larger neural network -- a teacher network \cite{cheng2017survey}. Such approaches can reduce the network size by about $5 \times$ with only a small drop in accuracy. These methods can also be combined with other model compression techniques such as network pruning and low-rank factorization methods.

In this paper, we consider a promising class of knowledge distillation methods where a new loss function is used to train the student network in addition to the cross-entropy loss so as to encourage the student features to more closely match the teacher features. Optimal transport is a principled method of comparing two distributions which may not share support \cite{villani2008optimal,peyre2019computational}.  We show that it is naturally applicable to the task of knowledge distillation and we design novel loss functions based on optimal transport tailored for the task of knowledge distillation. In our case, the two distributions are given by discrete measures of the student and teacher features, and the loss function encourages learning a student network that reduces the optimal transport cost between the two sets of features. We conduct experiments on image classification on a variety of teacher-student network pairs using standard datasets and show that optimal transport-based losses help improve student network accuracies better than other comparable losses which match teacher and student features.

\section{Background and related work}
Several classes of methods have been developed for model compression such as (a) pruning, where weights which do not affect the performance are removed \cite{WangEtAl2018,SrinivasBabu2015,HanEtAl2015}, (b) training compact CNNs with sparsity constraints \cite{LebedevLempitsky2016,ZhouEtAl2016,WenEtAl2016}, (c) low-rank approximations of convolutional filters and weights \cite{GongEtAl2014,WuEtAl2016,RastegariEtAl2016,HouEtAl2016} and (d) knowledge distillation where the performance of a smaller student network is improved by transferring the knowledge from a larger more accurate teacher network \cite{bucilua2006model,hinton2015distilling}. Please read this excellent survey by Cheng et al. \cite{cheng2017survey} for a more detailed treatment of these ideas. In this paper, we develop a novel knowledge distillation approach by devising loss functions based on the optimal transport cost between teacher and student feature distributions.

\subsection{Knowledge distillation methods}

The earliest methods of knowledge distillation (KD) were initially developed by Buciluǎ et al. \cite{bucilua2006model} and Hinton et al. \cite{hinton2015distilling} and has proven to be an effective method for transferring the knowledge learned by a complex teacher neural network into a simpler student network. Here, the basic idea is to use the softmax outputs of the last layer of the teacher network (i.e., soft labels) in the loss to train the student network, along with the usual cross-entropy loss to map to the one-hot vector given by the ground-truth label. Kullback-Liebler (KL) divergence, or a closely related variant, is used as the additional loss function between student outputs and the teacher's outputs, as both are probabilty distributions over the class labels and share support. In this paper, following recent works, we will refer to this loss as the KD loss. However, KD loss takes into account only the final outputs. Subsequent papers have looked at ways of exploiting intermediate layer outputs of the teacher network (i.e. from internal network layers) to further guide the training of the student network as in FitNets~\cite{romero2014fitnets}, attention transfer~\cite{zagoruyko2016paying}, relational knowledge distillation (RKD)~\cite{park2019relational}, probabilistic knowledge transfer \cite{passalis2018learning} etc. When combined with KD, all these methods result in significant improvements over KD alone. FitNets tries to match intermediate feature maps of teachers and students exactly, which may be too rigid of a constraint. On the other hand, RKD first computes pairwise relationships among teacher and student features in a batch and matches these geometric relationships between intermediate feature maps, rather than the feature maps themselves, which may be too loose of a constraint.  Attention transfer (AT) \cite{zagoruyko2016paying} matches intermediate attention maps which are derived from intermediate feature maps. 

In contrast to these methods, we propose a new loss for knowledge distillation based on optimal transport, to complement the standard KD loss. The central idea is to use optimal transport (OT) distance to measure the distance between the distributions of teacher and student feature maps in intermediate layers of the network. The OT loss function can be seen as a compromise between FitNets and RKD. The OT loss function is computed between two sets of features and is minimized when the two sets are equal \textbf{even if only up to a permutation}. Since we are matching distributions of feature maps rather than directly matching feature maps, this constraint is less stringent than that used by FitNets.  However, the constraint imposed by optimal transport is more stringent than the constraint imposed by RKD of merely matching pairwise geometric relationships computed separately in the teacher and student feature spaces.  Our conjecture is that a loss based on optimal transport provides a better balance as a way of encouraging student feature maps to be similar to teacher feature maps.  Our experiments will show that OT loss combined with KD loss improves over KD alone and also does better than FitNets + KD and RKD + KD. Another closely related work is Neuron Selectivity Transfer (NST) \cite{huang2017like} which employs Maximum Mean Discrepancy (MMD) to measure the distance between the teacher and student feature sets. MMD is an alternative class of geometry-aware distances between distributions and is computed using the distance between kernelized versions of the teacher and student distributions. However, MMD suffers both from theoretical and practical drawbacks \cite{feydy2019interpolating}. Theoretically, MMD does not faithfully capture the notion of distance in the original space. And practically, its gradient vanishes at the extreme points of the feature space. However, MMD is more easily scalable to large batches, compared to OT. We also note that recent methods also allow for interpolating between MMD and OT \cite{feydy2019interpolating}, however we do not investigate this direction in this paper.

A recent paper based on the idea of using contrastive objectives \cite{TianEtAl2019}  called Contrastive Representation Distillation \cite{tian2019contrastive} shows improvements over standard KD.  The basic idea is to use a contrastive loss \cite{TianEtAl2019} to compare features from internal layers between a teacher and a student network. For every pair of teacher and student features corresponding to the same input image, referred to as a "positive" pair, a large of set of "negative" teacher-student pairs is stored in memory. The authors then devise an algorithm to learn a student network that brings feature vectors in the positive pair closer to each other and drives feature vectors in negative student-teacher pairs away from each other.  It achieves state-of-the-art accuracy on CIFAR-100 using various combinations of student-teacher networks. To this end, the authors choose to maximize a lower bound of the mutual information between (a) the joint distribution of the student and teacher features and (b) the product of the student and teacher marginals. We will refer to this loss as CRD. We consider this idea to be complementary to both KD, as well as various other matching losses like FitNets, NST and OT. That is, in principle, contrastive versions of FitNets and OT can also be developed and we consider this to be part of our future work. In our experiments, we show that combining OT + KD + CRD improves the state of the art for knowledge distillation for CIFAR-100. We also show results for optimal transport knowledge distillation using ImageNet and SVHN where we achieve results competitive with or better than the current state of the art.

\subsection{Optimal transport for other applications in computer vision}

Optimal transport is a principled, theoretically well-studied problem and is applicable to tasks where the notion of distance between two distributions is important \cite{villani2008optimal}. The optimal transport problem was first studied in the late 18th century as a formalization of the problem of minimizing the cost of transporting resources to different locations. Around 1970, the concept of the Wasserstein distance (also known as the Earth Movers Distance) was proposed for a version of the optimal transport problem as a way of measuring the cost of transforming one probability distribution into another \cite{Dobrushin1970}.  As a way of comparing probability distributions, the Wasserstein distance does not require the two distributions to have overlapping support unlike KL-divergence, for example.  So in the case of comparing two distributions of feature vectors, if the distributions do not overlap, the KL-divergence would be infinity, while the Wasserstein distance would be a positive real number measuring the distance feature vectors in one distribution need to be moved to cover the feature vectors in the other.

In terms of applications to computer vision, optimal transport has previously been employed for domain adaptation, where the features of an unseen test domain are transformed into the training domain so as to achieve invariance to domain transformation \cite{courty2016optimal}. In a similar vein, it has been applied to the problem of style transfer where the low-level features of a content image are transformed in such a way that they more closely match those of a style image \cite{kolkin2019style}. Optimal transport loss functions have also been used to improve generative models as in the papers by Salimans et al. \cite{salimans2018improving} and Genevay et al. \cite{genevay2017learning}. We are the first to propose using optimal transport as a cost function for improved model compression.

\section{Optimal transport for knowledge distillation}
Given a trained teacher network and a student network which is to be trained using mini-batch gradient descent, we pass a training batch with $b$ images through both the networks. Let the teacher features in a batch be $X = \{\mathbf{x}^{(l)}_1, \mathbf{x}^{(l)}_2, \dots \mathbf{x}^{(l)}_b\}$ and the corresponding student features be $Y = \{\mathbf{y}^{(l)}_1, \mathbf{y}^{(l)}_2, \dots \mathbf{y}^{(l)}_b\}$ for some intermediate layer $l$. Let $\mathbf{c}$ and $\hat{\mathbf{c}}$ be the ground-truth class-conditional distributions for the batch (one-hot encodings) and the corresponding predictions by the student network. The student network is trained by minimizing a combined loss function given by 

\begin{equation}
     L = L_{CE}(\mathbf{c}, \hat{\mathbf{c}}) + \alpha \sum_{l=1}^L L_{OT}(X^{(l)}, Y^{(l)}),
     \label{eq:ot_loss}
\end{equation}

where $L_{CE}(\cdot, \cdot)$ is the usual cross-entropy loss used for classifcation and $L_{OT}(\cdot, \cdot)$ is the proposed optimal transport loss. In particular, we use the Wasserstein-1 metric, also called the Earth Movers Distance. $\alpha \in \mathbb{R}$ is used to balance the two losses. As we will discuss later, additional loss functions can be added to the above such as KD loss and CRD loss with appropriate weights. Note that, in order to convert the given discrete feature sets into distributions, we use a uniformly scaled Dirac (unit mass) measure at each point. The optimal transport cost is given by

\begin{align}
    L_{OT}(X^{(l)}, Y^{(l)}) = \min_{T \geq 0} \sum_{i,j} T^{(l)}_{i,j} C^{(l)}_{i,j} \quad
    \text{s.t.} \sum_{i} T_{i,j} = \sum_{j} T_{i,j} = \frac{1}{b},
    \label{ot}
\end{align}

where $T$ is called the transport matrix which encodes the soft-alignment between the two sets of vectors and $C$ is the cost matrix which contains the distance between all teacher-student feature pairs. In the case of model compression, we want to learn the student parameters such that the optimal transport cost is reduced. This means that the student features should be geometrically close to the teacher features. It is easy to see that this loss is more loose than FitNets~ \cite{romero2014fitnets} as it allows different examples fom the teacher features to be aligned to the student features, while it is more strict than RKD~\cite{park2019relational} as it requires the features to come closer to each other as measured by the distance function. In this paper, we employ the cosine distance on the unit sphere as the cost function. That is

\begin{equation}
\label{eq:cosine}
    C^{(l)}_{i,j} = 1 - \frac{\mathbf{x}^{(l) T}_i \mathbf{y}^{(l)}_j}{\norm{\mathbf{x}^{(l)}_i} \norm{\mathbf{y}^{(l)}_j}}
\end{equation}

The expression for $L_{OT}$ is computationally prohibitive to compute and scales cubically with $b$. Instead, we turn to two more tractable versions of the same expression, as discussed below.

\subsection{Inexact Proximal Optimal Transport (IPOT)}
A popular way of reducing the computational complexity of the OT problem is by adding an entropic regularization term \cite{cuturi2013sinkhorn}, which we call regularized OT (ROT): 

\begin{align}
    L_{ROT}(X^{(l)}, Y^{(l)}) = \min_{T \geq 0} \sum_{i,j} T^{(l)}_{i,j} C^{(l)}_{i,j} + \epsilon h(T) \quad
    \text{s.t.} \sum_{i} T_{i,j} = \sum_{j} T_{i,j} = \frac{1}{b},
    \label{ot}
\end{align}

where $h(T) = \sum_{i,j} T_{i,j} \log(T_{i,j})$ measures the entropy of the transportation matrix. The problem can then be solved efficiently using Sinkhorn iterations \cite{benamou2015iterative} which has been shown to scale close to quadratically with the batch size $b$ \cite{altschuler2017near}. However, the efficiency and numerical stability are sensitive to the choice of $\epsilon$ and in practice it is difficult to tune. Having too small a value $\epsilon$ leads to significantly larger number of Sinkhorn iterations needed for convergence, while having a large value of $\epsilon$ leads to numerical instability.

In this paper, we employ an improved version called the Inexact Proximal point method for exact Optimal Transport (IPOT) by Xie et al. \cite{xie2018fast} in order to compute $L_{OT}$ and henceforth refer to this method as IPOT shown in Algorithm \ref{algo:ipot}. Here, inexact proximal point iterations based on Bregman divergence are used. There exist works which show fast linear convergence under certain conditions for such methods. Xia et al. \cite{xie2018fast} show both theoretically and experimentally that the algorithm overcomes the drawbacks of earlier methods and converges to the exact solution with the same computational complexity as the Sinkhorn iterations. We note that it is easy to integrate algorithms to compute optimal transport with deep learning toolboxes with automatic differentiation and many dedicated libraries have also been developed for this purpose \cite{flamary2017pot,charlier2020kernel}. 

\begin{algorithm}[!htb]
\label{algo:ipot}
\begin{algorithmic}[1]
\caption{Algorithm to compute IPOT($\mathbf{\mu},\mathbf{\nu}, C$) \cite{xie2018fast}}
\label{alg:main}
\Input{The masses $\mathbf{\mu},\mathbf{\nu}$ at the feature points $X, Y$ and the cost matrix $C$ computed using Eqn. \eqref{eq:cosine}}
\Output{The optimal transport cost $L_{OT} = \sum_{i,j} T^*_{i,j} C_{i,j} $}
\Input{Initialize: $\mathbf{v} \leftarrow \frac{1}{b} \mathbf{1}_b, G_{i,j} = e^{-\frac{C_{i,j}}{\beta}}, T^{(0)} \leftarrow \mathbf{11}^T$}
\While{$t < N$}
    \State {$Q \leftarrow G \odot T^{(t)}$, where $\odot$ is the element-wise product}
    \State {$\mathbf{u} \leftarrow \frac{\mathbf{\mu}}{Q\mathbf{v}}, \mathbf{v} \leftarrow \frac{\mathbf{\nu}}{Q^T\mathbf{u}}$}
    \State {$T^{(t+1)} \leftarrow \text{diag}(\mathbf{u})Q\text{diag}(\mathbf{v})$}
    \EndWhile
\end{algorithmic}
\end{algorithm}

\subsection{Relaxed Earth-Mover's Distance (REMD)}
As we will see from the experiments, although IPOT performs well, the training time can be quite long, depending on the number of Sinkhorn iterations. Instead, we consider a computationally efficient relaxation of the optimal transport problem following the work of Kusner et al. \cite{kusner2015word}. Instead of solving Equation \eqref{ot}, two simpler problems are first generated by dropping one set of constraints. We then have

\begin{align}
    & R^{(1)}_{OT}(X^{(l)}, Y^{(l)}) = \min_{T \geq 0} \sum_{i,j} T^{(l)}_{i,j} C^{(l)}_{i,j} \quad \text{s.t.} \sum_{i} T_{i,j} = \frac{1}{b} \text{ and } \\
    & R^{(2)}_{OT}(X^{(l)}, Y^{(l)}) = \min_{T \geq 0} \sum_{i,j} T^{(l)}_{i,j} C^{(l)}_{i,j} \quad \text{s.t.} \sum_{j} T_{i,j} = \frac{1}{b}
    \label{ot}
\end{align}

The final relaxed EMD (REMD) is computed using

\begin{align}
    L_{REMD}(X^{(l)}, Y^{(l)}) & = \max(R^{(1)}_{OT}(X^{(l)}, Y^{(l)}), R^{(2)}_{OT}(X^{(l)}, Y^{(l)})) \\
                                & = \frac{1}{b} \max \left( \sum_{i} \min_j C^{(l)}_{i,j}, \sum_{j} \min_i C^{(l)}_{i,j} \right) 
\end{align}

\section{Experimental results}
\subsection{CIFAR-100 \cite{krizhevsky2009learning}}
This dataset consists of 50000 training images and 10000 test images. Each image needs to be classified into 1 of 100 fine-grained categories. Each class has about 600 images in the dataset. All the images are in RGB format and are of size $32 \times 32$.

We use the same teacher and student combinations as in the paper by Tian et al. \cite{tian2019contrastive} as they provide a common framework for benchmarking various knowledge distillation methods. The teacher and student networks on which we conduct experiments, the number of weights and the corresponding compression ratios are given in the appendix. 

\paragraph{Implementation details:} All the hyperparameters are held constant for all the methods, We use a batch size of 64 and train for $240$ epochs. An initial learning rate of $0.05$ is used and reduced by $\frac{1}{10}$ after $150, 180, 210$ epochs. We use Top-1 accuracy on the test set to compare the performance between various distillation methods. We compare our method with several other recent and popular knowledge distillation methods that also encourage similarity between teacher and student features at different layers including FitNets \cite{romero2014fitnets}, RKD \cite{park2019relational}, AT \cite{zagoruyko2016paying}, SP \cite{tung2019similarity}, CC \cite{peng2019correlation}, PKT \cite{passalis2018learning}, AB \cite{heo2019knowledge}, FT \cite{kim2018paraphrasing}, FSP \cite{yim2017gift} and NST \cite{huang2017like}, as well as KD \cite{hinton2015distilling} and CRD \cite{tian2019contrastive}. We report results on two methods of calculating the OT distance: REMD and IPOT and furthermore, the combination of the OT losses with KD and CRD. All the student networks are divided into 4 stages of approximately equal depth, independent of the architecture or depth of the overall network. The output of the fourth stage is the penultimate layer output, immediately before the softmax operation. During training of student networks, along with the softmax outputs required to calculate the cross-entropy loss, features at the outputs of these 4 intermediate stages for both student and teacher features are extracted. Then OT losses can be computed for one or more of four sets of features between the student and teacher which is also the procedure used for all the baseline algorithms. The number of stages used for loss computation is a hyperparameter. For all of our main experiments, we compute and add the the OT loss between all four sets. Later, as an ablative study, we show that even computing the OT loss at a single layer can yield close to optimal performance. Also note that when the teacher and student architectures are different as in the case of resnet32x4/ShuffleNetV2, we use additional embedding layers which map from the teacher space to the student space, following FitNets \cite{romero2014fitnets} and CRD \cite{tian2019contrastive}. Further details on this are provided in the appendix. When adding the OT loss term to the cross-entropy loss, we experimented with two weights for the OT loss (see Eq. \eqref{eq:ot_loss}), $\alpha = 0.9, 1.0$ and we report the better of the two results in the main paper. Accuracies obtained for both values are provided in the appendix. When using IPOT, we employ $\beta = 20$ following Xia et al. \cite{xie2018fast} and the number of proximal point iterations, $N = 50$, as it provides a good balance between training time and accuracy. In our experiments on an Nvidia GTX 1080, REMD and IPOT take about 2.5 hours and 10 hours, respectively, for training.

\paragraph{Results: } All the main results for CIFAR-100 are shown in Tables \ref{tbl:cifar100_same} and \ref{tbl:cifar100_combine}. The results for the baseline methods are reproduced from the paper by Tian et al. \cite{tian2019contrastive}. Table \ref{tbl:cifar100_same} contains the Top-1 accuracy on the CIFAR-100 test using various knowledge distillation methods for various pairs of teacher-student architecture pairs. We make the observation that both REMD and IPOT either outperform or yield similar performance to all comparable baselines with the exception of CRD. Note that CRD uses a contrastive objective, unlike all the other methods. More importantly, in Table \ref{tbl:cifar100_combine}, we show results when we combine various loss functions with the KD loss. Again, we see similar trends as before. However, the boosts in performance are more pronounced for both REMD+KD and IPOT+KD, which outperform all the baseline loss functions, including CRD+KD, for nearly all teacher-student pairs. We also show that we can obtain state-of-the-art results by combining IPOT+CRD+KD. For this setting, we use an equal weight of $1.0$ for all the four loss terms. 

\begin{table}[ht]

\setlength{\tabcolsep}{5pt}
\begin{center}
\begin{tabular}{lccccccc}
\toprule
\thead{Teacher \\ Student} & \thead{WRN-40-2 \\ WRN-16-2} & \thead{resnet56\\resnet20} & \thead{resnet110\\resnet20} & \thead{resnet110\\resnet32} & \thead{resnet32x4\\resnet8x4} & \thead{vgg13\\vgg8} & \thead{resnet32x4 \\ ShuffleNetV2}\\
\midrule
Teacher & 75.61 & 72.34 & 74.31 & 74.31 & 79.42 & 74.64 & 79.42\\
Student & 73.26 & 69.06 & 69.06 & 71.14 & 72.50 & 70.36 & 71.82 \\
\midrule

KD \cite{hinton2015distilling} & 74.92 & 70.66 & 70.67 & 73.08 & 73.33 & 72.98 & 74.45 \\

\midrule

CRD \cite{tian2019contrastive} & \textbf{75.48} & \textbf{71.16} & \textbf{71.46} & \textbf{73.48} & \textbf{75.51} & \textbf{73.94} & \textbf{75.65}\\
CRD+KD & 75.64 & \textcolor{red}{\textbf{71.63}} & \textcolor{red}{\textbf{71.56}} & \textcolor{red}{\textbf{73.75}} & 75.46 & 74.29 & 76.05\\

\midrule

FitNet \cite{romero2014fitnets} & 73.58 & 69.21 & 68.99 & 71.06 & 73.50 & 71.02 & 73.54\\
AT \cite{zagoruyko2016paying} & 74.08 & 70.55 & 70.22 & 72.31 & 73.44 & 71.43 & 72.73\\
SP \cite{tung2019similarity} & 73.83 & 69.67 & 70.04 & 72.69 & 72.94 & 72.68 & 74.56\\
CC \cite{peng2019correlation} & 73.56 & 69.63 & 69.48 & 71.48 & 72.97 & 70.71 & 71.29 \\
VID \cite{ahn2019variational} & 74.11 & 70.38 & 70.16 & 72.61 & 73.09 & 71.23 & 73.40\\
RKD \cite{park2019relational} & 73.35  & 69.61 & 69.25 & 71.82 & 71.90 & 71.48 & 73.21\\
PKT \cite{passalis2018learning} & 74.54 & 70.34 & 70.25 & 72.61 & 73.64 & 72.88 & 74.69\\
AB \cite{heo2019knowledge} & 72.50 & 69.47 & 69.53 & 70.98 & 73.17 & 70.94 & 74.31 \\
FT \cite{kim2018paraphrasing} & 73.25 & 69.84 & 70.22 & 72.37 & 72.86 & 70.58 & 72.50 \\
FSP \cite{yim2017gift} & 72.91 & n/a & 69.95 &  71.89 & 72.62 & 70.23 & n/a\\
NST \cite{huang2017like} & 73.68 & 69.60 & 69.53 & 71.96 & 73.30 & 71.53 & 74.68\\

\midrule

REMD & 74.19 & 70.66 & 70.76 & 72.96 & 73.97 & 72.45 & 73.58\\
REMD + KD & \textcolor{red}{\textbf{75.79}} & 71.59 & 70.98 & 73.66 & \textcolor{red}{\textbf{76.06}} & \textcolor{red}{\textbf{74.35}} & 76.66\\
IPOT & 74.79 & 71.04 & 70.79 & 72.87 & 74.19 & 72.80 & 73.97\\
IPOT + KD & 75.63 & 71.32 & 71.29 & {73.68} & 75.99 & 74.29 & \textcolor{red}{\textbf{76.78}}\\

\bottomrule

\end{tabular}
\vspace{5pt}
\caption{Comparison of various knowledge distillation methods on the CIFAR-100 dataset \cite{hinton2015distilling}. We are reporting the Top-1 accuracy ($\%$) on the test set. We observe that the OT loss functions proposed in this paper -- REMD and IPOT -- generally outperform or are very close to all comparable methods which also measure similarity between internal teacher and student features such as FitNets, RKD etc. We find that IPOT leads to slightly better performance compared to REMD. All the numbers except for the OT losses are retrieved from \cite{tian2019contrastive}. Although using a contrastive objective i.e., CRD yields better results than other loss functions, OT + KD leads to the best results in several cases. The best results without KD are shown in \textbf{bold} and the best results with KD are shown in \textcolor{red}{\textbf{red}}. In row 2, "Student" refers to the student networks trained without using a distillation loss. The numbers for the OT losses are averaged over 3 runs and the standard deviations are given in the appendix for better readability.}
\label{tbl:cifar100_same}
\end{center}
\vspace{-10pt}
\end{table}

\begin{table}[ht]

\setlength{\tabcolsep}{4.5pt}
\begin{center}
\begin{tabular}{lccccc}
\toprule
\thead{Teacher \\ Student} & 
\thead{WRN-40-2 \\ WRN-16-2} & 
\thead{resnet110\\resnet20} & 
\thead{resnet32x4\\resnet8x4} & 
\thead{vgg13\\vgg8} &
\thead{resnet32x4\\ShuffleNetV2}\\
\midrule
Teacher & 75.61 & 74.31 & 79.42 & 74.64 & 79.42 \\
Student (no distillation) & 73.26 & 69.06 & 72.50 & 70.36 & 71.82 \\
\midrule

KD \cite{hinton2015distilling} & 74.92 & 70.67 & 73.33 & 72.98 & 74.45 \\
\midrule
CRD+KD \cite{tian2019contrastive} & 75.64 & \textbf{71.56} & 75.46 & 74.29 & 76.05 \\
\midrule
FitNet+KD \cite{romero2014fitnets} & 75.12 & 70.67 & 74.66 & 73.22 & 75.15 \\
AT+KD \cite{zagoruyko2016paying} & 75.32 & 70.97 & 74.53 & 73.48 & 75.39 \\
SP+KD \cite{tung2019similarity} & 74.98 & 71.02 & 74.02 & 73.49 & 74.88 \\
CC+KD \cite{peng2019correlation} & 75.09 & 70.88 & 74.21 & 73.04 & 74.71 \\
VID+KD \cite{ahn2019variational} & 75.14 & 71.10 & 74.56 & 73.19 & 74.85 \\
RKD+KD \cite{park2019relational} & 74.89 & 70.77 & 73.79 & 72.97 & 74.55 \\
PKT+KD \cite{passalis2018learning} & 75.33 & 70.72 & 74.23 & 73.25 & 74.66 \\
AB+KD \cite{heo2019knowledge} & 70.27 & 70.97 & 74.40 & 73.35 & 74.99 \\
FT+KD \cite{kim2018paraphrasing} & 75.15 & 70.88 & 74.62 & 73.44 & 75.06 \\
NST+KD \cite{huang2017like} & 74.67 & 71.01 & 74.28 & 73.33 & 75.24 \\
\midrule
REMD + KD & \textbf{75.79} & 70.98 & \textbf{76.06} & \textbf{74.35} & 76.66\\
IPOT + KD & 75.63 & 71.29 & 75.99 & 74.29 & \textbf{76.78} \\
IPOT + CRD & 75.57 & 71.47 & 76.06 & 74.30 & 76.81 \\
IPOT + CRD + KD & \textcolor{red}{\textbf{76.22}} & \textcolor{red}{\textbf{71.81}} & \textcolor{red}{\textbf{76.82}} & \textcolor{red}{\textbf{74.79}} & \textcolor{red}{\textbf{76.81}} \\
\bottomrule
\end{tabular}
\vspace{5pt}
\caption{Comparison of various knowledge distillation methods on the CIFAR-100 dataset when combined with KD loss \cite{hinton2015distilling}. We are reporting the Top-1 accuracy ($\%$) on the test set. We can easily observe that the OT loss functions proposed in this paper -- REMD+KD and IPOT+KD -- outperform all comparable methods which measure similarity between internal teacher and student features. All the numbers except for the OT losses are retrieved from \cite{tian2019contrastive}. We also see that CRD + KD, which proposes a contrastive objective produces similar performance as REMD+KD and IPOT+KD. The best results are shown in  \textcolor{red}{\textbf{red}} and the second best results are shown in \textbf{bold}. In row 2, "Student" refers to the student networks trained with only the usual cross-entropy loss and without using a distillation loss. The numbers for the OT losses are averaged over 3 runs and the standard deviations are given in the appendix for better readability.}
\label{tbl:cifar100_combine}
\end{center}
\vspace{-10pt}
\end{table}

\paragraph{Ablation study:} In the experiments above, we use the student and teacher features from all four layers to compute the OT loss. Here, we find out their individual contribution to the final performance when using IPOT, as shown in Table \ref{tbl:ablation}. Surprisingly, we find that even though using all layers yields better performance, even computing the OT loss on a single layer yields nearly as good performance for this dataset. It is possible that combining specific pairs or triplets of layers rather than using all layers, may lead to even better performance. 

\begin{table}[]
    \centering
    \label{tbl:ablation}
    \begin{tabular}{ccc}
        \toprule
        Features used for loss computation & IPOT & IPOT + KD \\
        \midrule
        Output of stage 1 & 73.59 & 75.20 \\
        Output of stage 2 & 74.01 & 75.39 \\
        Output of stage 3 & 74.25 & 75.53 \\
        Output of stage 4 & 73.27 & 75.66 \\
        \midrule
        Using outputs of all 4 stages & 74.19 & 75.99 \\
        \bottomrule
    \end{tabular}
    \vspace{5pt}
    \caption{Contribution of IPOT loss to performance per layer used for loss computation. We report the Top-1 accuracy (\%) on the CIFAR-100 test set using resnet32x4 teacher and resnet8x4 student. Note that resnet8x4 without distillation yields 72.50\% accuracy.}
\end{table}

\subsection{Imagenet \cite{deng2009imagenet}} 

The ILSVRC2012 dataset, also called the ImageNet dataset is a large database containing 1.2 million RGB images in the training set and 50000 RGB images in the validation set. The task is image classification and each image belongs to one of $1000$ categories. As the test set is not available, the validation set is itself employed as the test set. All the images are resized to $224 \times 224$ for training. We conduct our experiment comparing the proposed optimal transport loss using IPOT with other loss functions which are applied at the intermediate layers including AT, SP and CC, as well as KD and CRD. A pretrained ResNet-34 released by PyTorch serves as the teacher network while a ResNet-18 is used as the student network.  The results are shown in Table \ref{tbl:imagenet} where we report the Top-1 and Top-5 error percentages on the validation set for all the methods. We see that IPOT loss leads to comparable performance with other loss functions while CRD clearly performs the best.  When IPOT is combined with KD, the top-1 error rate decreases to $28.88\%$ which is comparable to the $28.62\%$  top-1 error rate using CRD+KD.  Note that the all the results except those involving IPOT loss function are reported as in the paper by Tian et al. \cite{tian2019contrastive}.

\begin{table}[t]
\scriptsize
\setlength{\tabcolsep}{4.5pt}
\begin{center}
\begin{tabular}{l|cc|cccc|ccccc}
\toprule
 & Teacher & Student & KD & Online KD * & CRD & CRD+KD & AT & SP & CC & IPOT & IPOT+KD \\
\midrule
Top-1 & 26.69 & 30.25 & 29.34 & 29.45 & 28.83 & 28.62 & 29.30 & 29.38 & 30.04 & 29.54 & 28.88 \\
Top-5 & 8.58 & 10.93 & 10.12 & 10.41 & 9.87 & 9.51 & 10.00 & 10.20 & 10.83 & 10.48 & 9.66\\
\bottomrule

\end{tabular}

\caption{Top-1 and Top-5 error rates (\%) on ImageNet validation set for different knowledge distillation losses. The teacher network is ResNet-34 and the student network is ResNet-18. IPOT+KD performs close to the state-of-the art which is CRD+KD. "Student" refers to the student network trained with only the usual cross-entropy loss and without using a distillation loss.}

\vspace{-5pt}

\label{tbl:imagenet}
\end{center}
\end{table}

\subsection{The Street View House Numbers (SVHN) dataset \cite{netzer2011reading}}

SVHN contains images belonging to 10 classes consisting of 10 digits 0-9 cropped from real world images of house numbers. The images are of size $32 \times 32$ and have RGB channels. The training set contains 73257 images and the test set contains 26032 images. We perform a similar set of experiments as in the case of CIFAR-100. We use two pairs of teacher-student networks and for training the student networks, we use a batch size of $200$, an initial learning rate of $0.001$ and $150$ epochs for training the networks. The learning rate is reduced by $\frac{1}{10}$ after $80, 110, 135$ epochs. The remaining hyperparameters are identical to those employed for the experiments on CIFAR-100. We compare IPOT and REMD with KD, FitNets, RKD, PKT and CRD for two sets of teacher-student pairs. The results thus obtained are shown in Table \ref{tbl:svhn_results}. We immediately see that IPOT outperforms REMD in all cases. Overall, IPOT yields better results than other loss functions except CRD and RKD for this dataset. However, when KD is added to all the loss functions, IPOT+KD yields state-of-the-art performance for WRN-40-2/WRN-16-2 and the second-best results for resnet32x4/resnet8x4.

\begin{table}[t]
\scriptsize
\setlength{\tabcolsep}{1pt}
\begin{center}
\begin{tabular}{l|cc|c|cc|cc|cc|cc|cc|cc}
\toprule
 T-S pair & Teacher & Student & KD & CRD & CRD+KD & FitNet & Fitnet+KD & RKD & RKD+KD & PKT & PKT+KD & REMD & REMD+KD & IPOT & IPOT+KD \\
\midrule
\makecell{resnet32x4 \\ resnet8x4} & 94.36 & 90.39 & 94.49 & \textbf{94.96} & \textcolor{red}{\textbf{95.47}} & 91.32 & 94.48 & 93.30 & 94.58 & 90.77 & 94.38 & 89.66 & 94.49 & 91.63 & 94.73 \\
\midrule
\makecell{WRN-40-2 \\ WRN-16-2} & 94.52 & 93.45 & 95.22 & 94.74 & 95.25 & 93.93 & 95.27 & 95.23 & \textbf{95.39} & 93.68 & 95.15 & 93.15 & 94.94 & 94.28 & \textcolor{red}{\textbf{95.41}} \\
\bottomrule

\end{tabular}

\vspace{ 5pt}
\caption{Accuracy (\%) on the SVHN test set for different knowledge distillation losses for two teacher-student (T-S) pairs. IPOT+KD yields state-of-the-art performance for WRN-40-2/WRN-16-2 and the second-best results for resnet32x4/resnet8x4. The best results are shown in  \textcolor{red}{\textbf{red}} and the second best results are shown in \textbf{bold}. "Student" refers to the student networks trained with only the usual cross-entropy loss and without using a distillation loss.}

\vspace{-5pt}

\label{tbl:svhn_results}
\end{center}
\end{table}

\section{Discussion}
In this paper, we propose novel loss functions for model compression based on optimal transport. The loss functions measure the distance between the distribution of features from a teacher network and the distribution of student features. Minimizing these losses while training student networks encourages them to output features that belong to the same distribution as the teacher. We verify experimentally on CIFAR-100, ImageNet and SVHN that the proposed loss functions perform better or as well as various other loss functions proposed in the knowledge distillation literature. This is true across many teacher-student architecture pairs as well. As part of future work, developing ``contrastive" optimal transport loss functions for knowledge distillation, following the work of \cite{TianEtAl2019}, is a promising direction. The loss functions can encourage bringing closer teacher and student distributions for positive pairs and push farther apart distributions for negative pairs. These losses could further take into account the semantic content in the training images while creating positive and negative pairs. (CRD only labels a teacher/student feature pair as positive if it come from the same image, as opposed to coming from the same object class.) Another interesting avenue for future research involves developing and employing faster methods for comparing distributions such as using interpolated distances between MMD and OT which combines advantages of both methods \cite{feydy2019interpolating}.

\bibliographystyle{ieee}
\bibliography{references}

\appendix
\section{Number of parameters in the teacher and student networks used in the paper}

\begin{table}[h]
    \centering
    \label{tbl:teacher_student_parameters}
    \begin{tabular}{c|cc|cc|c}
        \toprule
        Datasets & \makecell{Teacher \\ architectures} & \makecell{\# weights \\ ($\times 10^6$)} & \makecell{Student \\ architectures} & \makecell{\# weights \\ ($\times 10^6$)} & \% reduction in \# weights \\
        \midrule
        \multirow{5}{*}{\makecell{CIFAR-100 / \\ SVHN}} & resnet110 & 1.74 & \multirow{2}{*}{resnet20} & \multirow{2}{*}{0.28} & 83.97 \\
        & resnet56 & 0.86 & & & 70.00 \\
        & resnet32x4 & 7.43 & resnet8x4 & 1.23 & 83.41\\
        & WRN-40-2 & 2.25 & WRN-16-2 & 0.70 & 68.81 \\
        & vgg13 & 9.46 & vgg8 & 3.97 & 58.10\\
        \midrule
        ImageNet & ResNet-34 & 21.28 & ResNet-18 & 11.17 & 48.00\\
        \bottomrule
    \end{tabular}
    \vspace{5pt}
    \caption{The number of parameters in the networks used in this paper and the compression achieved for every teacher-student pair.}
\end{table}

\section{Effect of the $\alpha$ and $\gamma$ in the IPOT-based loss functions}
The loss function used to train the student networks is given in Equation \eqref{eq:loss}. It shows that, in addition to the usual cross-entropy loss $L_{CE}(\cdot, \cdot)$ (between the ground-truth labels written as a one-hot encoding $\mathbf{c}$ and the output of the student network $\hat{\mathbf{c}}$), we add the optimal transport (OT) loss proposed in this paper, weighted by $\alpha$, and the KD loss \cite{hinton2015distilling} weighted by $\beta$. For the experiment on CIFAR-100, we investigate the effect of $\alpha$ and $\gamma$, and the results for different teacher-student pairs are shown in Table \ref{tab:supp2}. All the numbers show the top-1 accuracy on the test set averaged over 3 runs. Additionally, the table also shows the standard deviation for the 3 runs. 

\begin{equation}
\label{eq:loss}
     L = L_{CE}(\mathbf{c}, \hat{\mathbf{c}}_S) + \alpha \sum_{l=1}^L L_{OT}(X^{(l)}, Y^{(l)}) + \gamma L_{KD}(\hat{\mathbf{c}}_S, \hat{\mathbf{c}}_T),
\end{equation}

\begin{table}[]
    \centering
    \tiny
    \renewcommand{\tabcolsep}{1mm}
    \begin{tabular}{lccccccc}
    \toprule
    \backslashbox{Weights}{\thead{Teacher \\ Student}} & \thead{WRN-40-2 \\ WRN-16-2} & \thead{resnet56\\resnet20} & \thead{resnet110\\resnet20} & \thead{resnet110\\resnet32} & \thead{resnet32x4\\resnet8x4} & \thead{vgg13\\vgg8} & \thead{resnet32x4\\ShuffleNetV2}\\
    \midrule
         $\alpha=0.9, \gamma=0.0$ & $74.17 \pm 0.62$ & $71.04 \pm 0.32$ & $70.79 \pm 0.39$ & $72.87 \pm 0.06$ & $74.19 \pm 0.50$ & $72.75 \pm 0.15$ & $73.06 \pm$ 0.20\\
         $\alpha=1.0, \gamma=0.0$ & $74.79 \pm 0.09$ & $70.64 \pm 0.13$ & $70.67 \pm 0.05$ & $72.66 \pm 0.07$ & $74.03 \pm 0.18$ & $72.80 \pm 0.29$ & $73.97 \pm$ 0.62\\
         $\alpha=0.9, \gamma=1.0$ & $75.63 \pm 0.21$ & $71.21 \pm 0.28$ & $71.29 \pm 0.24$ & $73.54 \pm 0.25$ & $75.99 \pm 0.04$ & $74.29 \pm 0.31$ & $76.78\pm$ 0.12\\
         $\alpha=1.0, \gamma=1.0$ & $75.35 \pm 0.20$ & $71.32 \pm 0.22$ & $71.18 \pm 0.54$ & $73.68 \pm 0.57$ & $75.88 \pm 0.41$ & $74.21 \pm 0.36$ & $76.37 \pm$ 0.51\\
         \bottomrule
    \end{tabular}
    \vspace{5pt}
    \caption{Effect of $\alpha$ and $\gamma$ on the Top-1 accuracy for CIFAR-100 classification for IPOT and KD, and the corresponding standard deviation over 3 runs.}
    \label{tab:supp2}
\end{table}

\section{Details for different teacher-student architectures}
When the teacher and student features being compared come from different architectures and have different numbers of dimensions, the OT loss cannot be applied directly. This is also true for methods like FitNets \cite{romero2014fitnets} and CRD \cite{tian2019contrastive}. We follow the same protocol as these papers for making the dimensionalities of both the teacher and student features equal. In particular, if the features being compared are from the penultimate layer, we use separate linear layers to map both the teacher and student features to a common dimensional space. When the features being compared are derived from intermediate convolutional layers, we use additional convolutional layers to map the teacher features to the same dimensionality and shape as the student features. In our experiment with resnet32x4/ShuffleNetV2, the embedding dimensionality for the penultimate layer is set to $128$.

\end{document}